\documentclass{article}
\usepackage{ijcai17}

\usepackage{times}
\usepackage{bm}
\usepackage{makecell}
\usepackage{amssymb}
\usepackage{url}  
\usepackage{graphicx}  
\usepackage{epstopdf}
\usepackage{subfigure}
\usepackage{algorithm}
\usepackage{algorithmic}
\usepackage{array}
\hfuzz=\maxdimen
\title{Joint Adaptive Neighbours and Metric Learning\\ for Multi-view Subspace Clustering}
\author{Nan Xu$^1$, Yanqing~Guo$^1$, Jiujun Wang$^1$, Xiangyang Luo$^2$, and Ran He$^3$$^,$$^4$\\
$^1$School of Information and Communication Engineering, Dalian University of Technology\\
$^2$State Key Laboratory of Mathematical Engineering and Advanced Computing,\\
Zhengzhou Science and Technology Institute\\
$^3$National Laboratory of Pattern Recognition, CASIA;
$^4$University of Chinese Academy of Sciences (UCAS)\\
\{{xunan, jiujunwang}\}@mail.dlut.edu.cn, guoyq@dlut.edu.cn, xiangyangluo@126.com, rhe@nlpr.ia.ac.cn}

\begin{document}

\maketitle

\begin{abstract}
 Due to the existence of various views or representations in many real-world data, multi-view learning has drawn much attention recently. Multi-view spectral clustering methods based on similarity matrixes or graphs are
pretty popular. Generally, these algorithms learn informative graphs by directly utilizing original data. However, in the real-world applications, original data often contain noises and outliers that lead to unreliable graphs. In addition, different views may have different contributions to data clustering. In this paper, a novel Multi-view Subspace Clustering method unifying Adaptive neighbours and Metric learning (MSCAM), is proposed to address the above problems. In this method, we use the subspace representations of different views to adaptively learn a consensus similarity matrix, uncovering the subspace structure and avoiding noisy nature of original data. For all views, we also learn different Mahalanobis matrixes that parameterize the squared distances and consider the contributions of different views. Further, we constrain the graph constructed by the similarity matrix to have exact $c$ ($c$ is the number of clusters) connected components. An iterative algorithm is developed to solve this optimization problem. Moreover, experiments on a synthetic dataset and different real-world datasets demonstrate the effectiveness of MSCAM.
\end{abstract}

\section{Introduction}

In recent years, learning multi-view data has increasingly attracted research attention in many real-world applications, because data represented by different features or collected from different sources are very common. For instance, documents can have different languages; web pages can be described by different characteristics, e.g., hyperlinks and texts; images can have many descriptions with respect to different kinds of features like color or texture features. Different views or features can capture distinct perspectives of data that are complementary to each other. Thus, how to integrate these heterogeneous features and uncover the underlying structure of data is a critical problem for multi-view learning. In this paper, we focus on an unsupervised scenario, i.e., multi-view spectral clustering.

In the past decades, various spectral clustering algorithms have been proposed \cite{shi2000normalized,ng2002spectral,zelnik2005self,von2007tutorial,nie2014clustering,chang2015convex}. These methods can achieve promising clustering performance for an individual view. However, multiple views containing different information can describe the data more accurately and improve the clustering performance. \cite{zhou2007spectral} generalize the single-view spectral clustering normalized cut to the multi-view case. \cite{blaschko2008correlational} introduce the Canonical Correlation Analysis (CCA) to map multi-view data into a low-dimensional subspace. There are also some methods using co-training or co-regularization strategies to integrate different information of views \cite{kumar2011coa,kumar2011co}. In addition, \cite{cai2011heterogeneous} integrate heterogeneous features to learn a shared Laplacian matrix and improve model robustness with a non-negative constraint. \cite{wang2014multi} utilize the minimax optimization to obtain a universal feature embedding and a consensus clustering result. \cite{nie2017multi} simultaneously perform local structure learning and multi-view clustering in which the weight is automatically determined for each view. Recently, self-representation subspace based multi-view spectral clustering methods have been developed due to the effectiveness \cite{cao2015diversity,gao2015multi,yin2015multi,zhang2015low,wang2017exclusivity}. These methods aim to discover underlying subspaces embedded in original data for clustering accurately.

Although the previous multi-view spectral clustering methods can achieve promising performance, there still exist drawbacks. First, spectral methods need the high-quality similarity matrix. The previous methods directly learn the similarity matrix utilizing original data. However, in real-world datasets, data often contain noises and outliers, thus the similarity matrix learned from original data is unreliable. Second, different views have different contributions to data clustering. The previous methods use the Euclidean metric to learn the similarity matrix. For given data, the Euclidean
distance among them is fixed, which cannot consider different contributions of views. Finally, for multi-view spectral clustering, the \emph{k}-means procedure in spectral clustering requires the strict initialization, which influences the final clustering performance \cite{ng2002spectral}.

In this paper, we propose a novel subspace based multi-view spectral clustering method, named Multi-view Subspace Clustering unifying Adaptive neighbours and Metric learning (MSCAM) to address the aforementioned problems. In this method, we learn the subspace representations of original data for each view. By utilizing these subspace representations to adaptively learn a consensus similarity matrix, we can alleviate the influence of noises and outliers. Meanwhile, for each view, we learn the most suitable Mahalanobis matrix to parameterize the squared distance. The motivation is that due to the complexity of noises and outliers, different views have different contributions to clustering data. Thus we propose to use Mahalanobis metric to dynamically rescale data of each view. Different Mahalanobis matrixes are learned to weigh different contributions of views. Finally, we constrain the graph constructed by the similarity matrix to have exact $c$ connected components. Here, the number of clusters is $c$. In this way, the learned graph can be employed to cluster directly without the \emph{k}-means procedure.

The main contributions of our work are as follows:

\begin{itemize}
\item We adaptively learn a consensus similarity matrix in the subspace rather than the original space that may have noises and outliers.
\item Mahalanobis metric is employed to parameterize the squared distance of each view, which considers the contributions of different views to data clustering compared with the Euclidean metric.
\item We add a constraint on the graph constructed by the similarity matrix to replace the \emph{k}-means procedure.
\item Extensive comparison experiments demonstrate that our MSCAM method outperforms other state-of-the-art multi-view clustering approaches.
\end{itemize}

\section{Related Work}
\subsection{Notation Summary}
Lowercase letters $\left( {m,n,...} \right)$ denote scalars while bold lowercase letters $\left( \bm{m,n,...} \right)$ denote vectors. Bold uppercase letters $\left( \bm{M,N,...} \right)$ mean matrixes. For an arbitrary matrix $\bm{M}$, ${\bm{m}_i}$ means the ${i^{th}}$ column of $ \bm{M} $ and ${m_{ij}}$ stands for the ${j^{th}}$ element in ${\bm{m}_i}$.  ${\bm{M}^T}$ and $tr(\bm{M})$ denote the transpose and trace of $\bm{M}$, respectively. $\left\|  \cdot  \right\|$ and ${\left\| \cdot \right\|_F}$ represent the ${l_2}$ norm and Frobenius norm, respectively. For two matrixes with the same size, $\left\langle \bm{M,N} \right\rangle$ represents the inner product. Moreover, $\bm{1}$ and $\bm{I}$ represent the vectors of all ones and identity matrix with proper sizes, respectively.

\subsection{Adaptive Neighbours Clustering}
For clustering tasks, the local correlation of original data plays an important role. Recently, many clustering methods considering the local correlation have been developed \cite{nie2014clustering,guo2015robust,nie2016constrained,zhao2016incomplete}. Let $\bm{X} = \left[ {{\bm{x}_1},{\bm{x}_2},...,{\bm{x}_n}} \right] \in {\Re ^{d \times n}}$ be the data matrix with $n$ data points, where $d$ is the dimension of features. The Euclidean (squared) distance is used as a measure to decide the \emph{k}-nearest data of each data point. For each data point ${\bm{x}_i}$, all data points can be the neighbour of ${\bm{x}_i}$ with the probability ${a_{ij}}$. Generally, a smaller distance $\left\| {{\bm{x}_i} - {\bm{x}_j}} \right\|_2^2$ indicates that a larger probability ${a_{ij}}$ should be allocated. Therefore, the probabilities $\left. {{a_{ij}}} \right|_{j = 1}^n$ can be determined by solving the following problem
\begin{equation}\mathop {\min }\limits_{\bm{a}_i^T\bm{1} = 1,0 \le {a_{ij}} \le 1} \sum\limits_{j = 1}^n {\left\| {{\bm{x}_i} - {\bm{x}_j}} \right\|_2^2{a_{ij}}},
\end{equation}
where ${\bm{a}_i} \in {\Re ^{n \times 1}}$  is a vector, whose ${j^{th}}$ element is ${a_{ij}}$.

For problem (1), to rule out the trivial solution, where only the nearest data point of ${\bm{x}_i}$ is assigned probability 1 and the similarity of all the other points would be probability 0 , a penalty term is added to constrain the probability ${a_{ij}}$. For all the data points, the model with adaptive neighbours is
\begin{equation}\begin{array}{cl}\mathop {\min }\limits_{\bm{A}} &{\sum\limits_{i,j = 1}^n {\left( {\left\| {{\bm{x}_i} - {\bm{x}_j}} \right\|_2^2{a_{ij}} + \gamma a_{ij}^2} \right)} }\\[3mm]
s.t.&{\forall i,\bm{a}_i^T\bm{1} = 1,0 \le {a_{ij}} \le 1},
\end{array}
\end{equation}
where $\gamma  > 0$ is the trade-off parameter. After obtaining the similarity matrix $\bm{A}$ , the spectral clustering \cite{ng2002spectral} can be performed to get final clustering results.

\subsection{Multi-view Subspace Clustering (MSC)}
Subspace clustering aims to obtain a similarity matrix in the learned underlying subspace of original data and perform spectral clustering \cite{lu2012robust,elhamifar2013sparse,liu2013robust}.

In the dataset, each data point can be reconstructed by an effective combination of other points, i.e., $\bm{X} = \bm{XZ} + \bm{E}$, where $\bm{Z} = \left[ {{\bm{z}_1},{\bm{z}_{2}},...,{\bm{z}_n}} \right] \in {\Re ^{n \times n}}$ is the subspace representation matrix and $\bm{E} \in {\Re ^{d \times n}}$ is the error matrix.

Usually, the subspace clustering model can be written in the following form
\begin{equation}\begin{array}{cl}
\mathop {\min }\limits_{\bm{Z},\bm{E}} &{F(\bm{X},\bm{XZ}) + \alpha \Psi (\bm{Z})} \\[2mm]
s.t.&{{\rm{ }}\bm{X} = \bm{XZ} + \bm{E},}
\end{array}
\end{equation}
where $F(\bm{X},\bm{XZ})$ and $\Psi (\bm{Z})$ denote the error loss term and the regularized term, respectively. $\alpha$ is the trade-off parameter. We can obtain the subspace representation $\bm{Z}$ where the nonzero elements mean that the corresponding data points are from the same subspace. Then the similarity matrix $\bm{A} = {{\left( {\left| \bm{Z} \right| + \left| {{\bm{Z}^T}} \right|} \right)} \mathord{\left/
 {\vphantom {{\left( {\left| \bm{Z} \right| + \left| {{\bm{Z}^T}} \right|} \right)} 2}} \right.
 \kern-\nulldelimiterspace} 2}$ can be constructed. Afterwards, the spectral clustering \cite{ng2002spectral} is performed on the similarity matrix $\bm{A}$ to obtain clustering results.

Recently, subspace clustering is extended to the multi-view setting because of its effectiveness. Generally, these multi-view subspace clustering models can be formulated as
\begin{equation}\begin{array}{cl}
\mathop {\min }\limits_{{\bm{Z}_v},{\bm{E}_v}} &{\sum\limits_{v = 1}^V {F\left( {{\bm{X}_v},{\bm{X}_v}{\bm{Z}_v}} \right) + \alpha \Psi \left( {{\bm{Z}_v}} \right)}} \\[3mm]
s.t.&{{\rm{   }}\forall v,{\bm{X}_v} = {\bm{X}_v}{\bm{Z}_v} + {\bm{E}_v},}
\end{array}
\end{equation}
where ${\bm{X}_v} \in {\Re ^{{d^v} \times n}}$, ${\bm{Z}_v} \in {\Re ^{n \times n}}$, and ${\bm{E}_v} \in {\Re ^{{d^v} \times n}}$ (${d^v}$ is the dimension of features for the ${v^{th}}$ view) denote the data matrix, the subspace representation and the error matrix for the ${v^{th}}$ view ($v = 1,2,...,V$, $V$ is the number of views), respectively. However, these spectral clustering methods construct graphs from original data that are corrupted and require strict initialization, which results in suboptimal results. In contrast, we construct the graph in the underlying subspace of original data. We also learn different Mahalanobis matrixes considering the contributions of different views and remove the \emph{k}-means procedure from spectral clustering.

\section{Methodology}
In this section, we present the proposed MSCAM method, which utilizes subspace representations and Mahalanobis metric to adaptively learn a consensus similarity matrix. In addition, we add a constraint on the constructed graph to optimize spectral clustering.
\subsection{MSC with Adaptive Neighbours}
For multi-view data, ${\bm{X}_v}$ denotes the original data of the ${v^{th}}$ view. We extend the basic subspace clustering model to multi-view domains as follows.
\begin{equation}\mathop {\min }\limits_{{\bm{Z}_v}} \sum\limits_v {\left( {\left\| {{\bm{X}_v} - {\bm{X}_v}{\bm{Z}_v}} \right\|_F^2 + \alpha \left\| {{\bm{Z}_v}} \right\|_F^2} \right)} ,
\end{equation}
where $\alpha  > 0$ is the regularization parameter. ${\bm{Z}_v}$ is the learned subspace representation for each view. In our objective function, the Frobenius norm constraint on ${\bm{Z}_v}$ can improve the model robustness according to \cite{lu2012robust}.

Adaptive neighbours explore the local correlation of original data to improve the clustering performance. Moreover, subspace representations can uncover the underlying subspace structure of the original data and alleviate the influence of noises and outliers in original data. Therefore, we utilize the subspace representations rather than original data to learn a consensus similarity matrix for all views as
\begin{equation}\begin{array}{cl}
\mathop {\min }\limits_{{\bm{Z}_v},\bm{A}} &{\sum\limits_v {\left( {\left\| {{\bm{X}_v} - {\bm{X}_v}{\bm{Z}_v}} \right\|_F^2 + \alpha \left\| {{\bm{Z}_v}} \right\|_F^2} \right)} }\\[3mm]
&{+ \lambda \sum\limits_v {\sum\limits_{i,j} {\left\| {\bm{z}_i^v - \bm{z}_j^v} \right\|_2^2} {a_{ij}}}  + \gamma \left\| \bm{A} \right\|_F^2}\\[3mm]
s.t.&{{\rm{   }}\forall i,{\rm{ }}\bm{a}_i^T\bm{1} = 1,{\rm{ }}{a_{ij}} \ge 0,}
\end{array}
\end{equation}
where $\alpha$, $\lambda$, and $\gamma$ are three trade-off parameters. For the data point $\bm{x}_i$, $\bm{z}_i^v$ is the ${i^{th}}$ subspace representation in the ${v^{th}}$ view. For all views, we learn a consensus similarity matrix $\bm{A}$ with adaptive neighbours. Therefore, the data points in each view can be allocated to the most suitable cluster and the consistency of clustering results across views can be preserved. We name the method Multi-view Subspace Clustering with Adaptive Neighbours as MSCAN.

\subsection{Joint Adaptive Neighbours and Metric Learning}
According to MSCAN, we employ the subspace representations to adaptively learn a consensus similarity matrix, which alleviates the influence of noises and outliers in original data to some extent. However, due to the complexity of noises and outliers for multi-view data, different views may have different contributions to clustering. The MSCAN model utilizes the Euclidean distance as the metric to learn the similarity matrix. Thus for given subspace representations, the similarity among them will not change. That results in a suboptimal graph. By contrast, we use Mahalanobis metric to learn the similarity matrix for two reasons. For one thing, Mahalanobis metric aims to learn the Mahalanobis matrix that can parameterize the squared distance \cite{xing2003distance}. Moreover, learning the Mahalanobis matrix is equivalent to learning a rescaling of data, which indicates that the similarity among subspace representations will adaptively change until obtaining the satisfactory similarity matrix. For another, the Mahalanobis matrix can be decomposed into the form of matrix product, i.e., $\bm{P}{\bm{P}^T}$, which makes our model easy to solve. Our goal is to utilize the subspace representations to adaptively learn the similarity matrix. Therefore, considering the contributions of different views, we employ the Mahalanobis metric to rescale the subspace representations of different views, which changes the similarity relationships among subspace representations.

\subsubsection{Mahalanobis metric Learning} First, we introduce the Mahalanobis distance based metric learning.
Given the dataset $\left\{ {{\bm{x}_i}} \right\}_{i = 1}^n \subseteq {\Re ^d}$, consider learning the Mahalanobis metric of the form
\begin{equation}
{\left\| {{\bm{x}_i} - {\bm{x}_j}} \right\|_M} = \sqrt {{{({\bm{x}_i} - {\bm{x}_j})}^T}\bm{M}({\bm{x}_i} - {\bm{x}_j})} ,
\end{equation}
where $\bm{M}\ge 0$ is a positive semi-definite Mahalanobis matrix. Setting $\bm{M} = \bm{I}$, the formula (7) is the Euclidean distance; if let $\bm{M}$ be diagonal, it means that different weights are assigned to different axes in the metric learning; more usually, $\bm{M}$ can parameterize the squared distances. In addition, the Mahalanobis metric learning can rescale the data point ${\bm{x}_i}$ to ${\bm{M}^{1/2}}{\bm{x}_i}$. Then, a normal Euclidean metric learning can be employed in terms of these rescaled data points. The effectiveness of Mahalanobis metric learning has been verified in clustering \cite{xing2003distance} and classification \cite{weinberger2009distance,cao2013similarity}. In our paper, we employ the Mahalanobis distance as a metric since it can improve the clustering performance by well weighing the contributions of different views.

\subsubsection{Mahalanobis metric Induced MSC} Considering the contributions of different views, we learn different Mahalanobis matrixes in the underlying subspace for different views. Therefore, the Mahalanobis metric induced objective function can be formulated as
\begin{equation}\begin{array}{cl}
\mathop {\min }\limits_{{\bm{Z}_v},{\bm{M}_v}, \bm{A}} &{\sum\limits_v {\left( {\left\| {{\bm{X}_v} - {\bm{X}_v}{\bm{Z}_v}} \right\|_F^2 + \alpha \left\| {{\bm{Z}_v}} \right\|_F^2} \right)} }\\[3mm]
&{+ \lambda \sum\limits_v {\sum\limits_{i,j} {\left\| {\bm{z}_i^v - \bm{z}_j^v} \right\|_M^2} {a_{ij}}}  + \gamma \left\| \bm{A} \right\|_F^2}\\[3mm]
s.t.&{{\rm{   }}\forall i,{\rm{ }}\bm{a}_i^T\bm{1} = 1,{\rm{ }}{a_{ij}} \ge 0,}
\end{array}
\end{equation}
where $\left\| {\bm{z}_i^v - \bm{z}_j^v} \right\|_M^2 = {(\bm{z}_i^v - \bm{z}_j^v)^T}{\bm{M}_v}(\bm{z}_i^v - \bm{z}_j^v)$, and ${\bm{M}_v} \in {\Re ^{n \times n}}$ is a positive semi-definite Mahalanobis matrix. Usually, ${\bm{M}_v}$ can be decomposed as ${\bm{M}_v} = {\bm{P}_v}\bm{P}_v^T$, where ${\bm{P}_v} \in {\Re ^{n \times p}}$ and $p \le n$. In this way, Mahalanobis metric learning can be seen as finding a linear projection $\bm{P}_v$. Therefore, we can rewrite the model (8) as
\begin{equation}\begin{array}{cl}
\mathop {\min }\limits_{{\bm{Z}_v},{\bm{P}_v},\bm{A}} &{\sum\limits_v {\left( {\left\| {{\bm{X}_v} - {\bm{X}_v}{\bm{Z}_v}} \right\|_F^2 + \alpha \left\| {{\bm{Z}_v}} \right\|_F^2} \right)} }\\[3mm]
&{+ \lambda \sum\limits_v {\sum\limits_{i,j} {\left\| {\bm{P}_v^T(\bm{z}_i^v - \bm{z}_j^v)} \right\|_2^2} {a_{ij}}} + \gamma \left\| \bm{A} \right\|_F^2}\\[3mm]
s.t.&{{\rm{   }}\forall v,{\rm{ }}\bm{P}_v^T{\bm{Z}_v}\bm{Z}_v^T{\bm{P}_v} = {\bm{I}_p};{\rm{   }}\forall i,{\rm{ }}\bm{a}_i^T\bm{1} = 1,{\rm{ }}{a_{ij}} \ge 0}.
\end{array}
\end{equation}

In this model, the orthogonal constraint on ${\bm{P}_v}^T{\bm{Z}_v}\bm{Z}_v^T{\bm{P}_v}$ actually aims to learn a linear projection $\bm{P}_v$ to transfer original $n$-dimensional subspace representations into a $p$-dimensional uncorrelated space.
\begin{algorithm}[t] 
\caption{: Algorithm for solving the problem (\ref{admm})} 
\label{Algorithm 1} 
\begin{algorithmic}[1] 
\REQUIRE $\quad $\\Data matrix ${\bm{X} = \left\{ {{\bm{X}_1},{\bm{X}_2},...,{\bm{X}_v}} \right\}}$, ${\bm{X}_v} \in {\Re ^{{d^v} \times n}}$, parameters $\alpha$ and ${\lambda}$. 
\STATE Initialize ${\bm{Z}_0}$, ${\bm{P}_0}$, ${\bm{L}_0}= {\bm{D}_0} - {\bm{A}_0}$, ${\bm{Y}_0^{(1)}} = 0$, ${\bm{Y}_0^{(2)}} = 0$, $\varepsilon  = {10^{ - 6}}$, $\rho  = 1.1$, $t = 0$;
\REPEAT
\STATE Update ${\bm{Z}_t}$, ${\bm{P}_t}$, ${\bm{M}_t}$, and ${\bm{W}_t}$;

\STATE Update $\bm{Y}_t^{(1)}$, $\bm{Y}_t^{(2)}$;
\STATE Update ${\mu _{t + 1}} \leftarrow \rho {\mu _t}$;

\STATE $t = t + 1$;
\UNTIL{convergence};
\ENSURE{${\bm{Z}_{t+1}}$, ${\bm{P}_{t+1}}$.}
\end{algorithmic}
\end{algorithm}

\begin{algorithm}[t] 
\caption{: Algorithm for solving the problem (\ref{11})} 
\label{Algorithm 2} 
\begin{algorithmic}[1] 
\REQUIRE $\quad $\\Data matrix ${\bm{X} = \left\{ {{\bm{X}_1},{\bm{X}_2},...,{\bm{X}_v}} \right\}}$, parameters $\alpha$ and ${\lambda}$, number of classes $c$. 
\STATE Initialize ${\bm{A}}$;
\REPEAT
\STATE Given ${\bm{A}}$, update ${\bm{Z}_v}$ and ${\bm{P}_v}$ by solving problem (\ref{admm}) via Algorithm \ref{Algorithm 1};

\STATE Given ${\bm{Z}_v}$ and ${\bm{P}_v}$, update ${\bm{A}}$ by solving problem (\ref{15});

\UNTIL{convergence};
\ENSURE{${\bm{A}}$.}
\end{algorithmic}
\end{algorithm}
\subsection{Graph Constraint Term}
After obtaining the similarity matrix $\bm{A}$, a spectral clustering method \cite{ng2002spectral} is performed to get final clustering results. However, due to the \emph{k}-means procedure, spectral clustering depends on the initialization that influences the final clustering performance.

Inspired by \cite{nie2017multi}, the graph obtained via $\bm{A}$ shares exact $c$ connected components. As a result, it yields explicit clustering results with the similarity matrix $\bm{A}$. We first introduce Theorem 1 as follows.

\vspace{0.2cm}
\noindent\textbf{Theorem 1.} \textit{In the graph obtained by the similarity matrix $\bm{A}$, the number of connected components is equal to the multiplicity $c$ of 0 as the eigenvalue of the (nonnegative) Laplacian matrix $\bm{L}$.}\vspace{0.2cm}

Therefore, the graph with $c$ connected components means that the Laplacian matrix $\bm{L}$ ($\bm{L} = \bm{D} - \bm{A}$, where ${\bm{D}_{ii}} = \sum\nolimits_j {{a_{ij}}}$) should have $c$ zero eigenvalues. According to the theorem in \cite{fan1949theorem}, we let the dimension of the projection $\bm{P}_v$ be $c$ ($p = c$) and have
\begin{equation}\begin{array}{cl}\mathop {\min }\limits_{{\bm{P}_v}} &{tr(\bm{P}_v^T{\bm{Z}_v}\bm{LZ}_v^T{\bm{P}_v}) }\\ [3mm] s.t.&{\bm{P}_v^T{\bm{Z}_v}\bm{Z}_v^T{\bm{P}_v} = {\bm{I}_c},}
\end{array}
\end{equation}
where ${\bm{P}_v} \in {\Re ^{n \times c}}$ and ${\bm{I}_c} \in {\Re ^{c \times c}}$. Hence, we obtain our final multi-view subspace clustering objective function
\begin{equation}\begin{array}{cl}
\mathop {\min }\limits_{{\bm{Z}_v},{\bm{P}_v},\bm{A}} &{\sum\limits_v {\left( {\left\| {{\bm{X}_v} - {\bm{X}_v}{\bm{Z}_v}} \right\|_F^2 + \alpha \left\| {{\bm{Z}_v}} \right\|_F^2} \right)} }\\[3mm]
&{+ \lambda \sum\limits_v {tr(\bm{P}_v^T{\bm{Z}_v}\bm{LZ}_v^T{\bm{P}_v})} + \gamma \left\| \bm{A} \right\|_F^2}\\[3mm]
s.t.&{{\rm{   }}\forall v,{\rm{ }}\bm{P}_v^T{\bm{Z}_v}\bm{Z}_v^T{\bm{P}_v} = {\bm{I}_c};{\rm{  }}\forall i,{\rm{ }}\bm{a}_i^T\bm{1} = 1,{\rm{ }}{a_{ij}} \ge 0.}
\end{array}\label{11}
\end{equation}

The graph constraint term leads to explicit clustering results without the \emph{k}-means procedure, which improves the final clustering performance.

\section{Optimization}
The model (11) is not jointly convex, thus we solve the model (11) by iteratively optimizing each variable (${\bm{Z}_v}$, ${\bm{P}_v}$, and $\bm{A}$) while fixing other variables. In addition, we offer the convergence analysis.

\subsection{Optimization Procedure}
\subsubsection{Fixing $\bm{A}$, Update ${\bm{Z}_v}$ and ${\bm{P}_v}$}
When $\bm{A}$ is fixed, the model (11) can be formulated as
\begin{equation}\begin{array}{cl}
\mathop {\min }\limits_{{\bm{Z}_v},{\bm{P}_v}} &{\sum\limits_v {\left( {\left\| {{\bm{X}_v} - {\bm{X}_v}{\bm{Z}_v}} \right\|_F^2 + \alpha \left\| {{\bm{Z}_v}} \right\|_F^2} \right)} }\\[3mm]
&{+ \lambda \sum\limits_v {tr(\bm{P}_v^T{\bm{Z}_v}\bm{LZ}_v^T{\bm{P}_v})}}\\[3mm]
s.t.&{{\rm{   }}\forall v,{\rm{ }}\bm{P}_v^T{\bm{Z}_v}\bm{Z}_v^T{\bm{P}_v} = {\bm{I}_c}.}
\end{array}\label{12}
\end{equation}

For convenience, we ignore the subscript tentatively and rewrite the above formula as
\begin{equation}\begin{array}{cl}
\mathop {\min }\limits_{{\bm{Z}},{\bm{P}}} &{ { {\left\| {{\bm{X}} - {\bm{X}}{\bm{Z}}} \right\|_F^2 + \alpha \left\| {{\bm{Z}}} \right\|_F^2} } }+ \lambda \hspace{0.05cm}tr({\bm{P}^T}\bm{ZL}{\bm{Z}^T}\bm{P})\\[3mm]
s.t.&{{\rm{   }}{\rm{ }}\bm{P}^T{\bm{Z}}\bm{Z}^T{\bm{P}} = {\bm{I}_c}.}
\end{array}\label{admm}
\end{equation}

To minimize the problem (13), we set ${\bm{Z}^T}\bm{P} - \bm{W} = 0$ and $\bm{P} - \bm{M} = 0$, then utilize the Alternating Direction Method of Multipliers (ADMM) algorithm \cite{boyd2011distributed} to optimize. The augmented Lagrangian function is given by
\begin{equation}\begin{array}{l}
{\cal L}(\bm{Z},\bm{P},\bm{M},\bm{W},{\bm{Y}^{(1)}},{\bm{Y}^{(2)}})\\[3mm]
 = \left\| {\bm{X} - \bm{XZ}} \right\|_F^2 + \alpha \left\| \bm{Z} \right\|_F^2 + \lambda \hspace{0.05cm}tr({\bm{W}^T}\bm{LW})\\[3mm]
 + \left\langle {{\bm{Y}^{(1)}},\bm{W} - {\bm{Z}^T}\bm{P}} \right\rangle  + \left\langle {{\bm{Y}^{(2)}},\bm{M} - \bm{P}} \right\rangle \\[3mm]
 + \frac{\mu }{2}( {\left\| {\bm{W} - {\bm{Z}^T}\bm{P}} \right\|_F^2 + \left\| {\bm{M} - \bm{P}} \right\|_F^2} )\\[3mm]
s.t.\hspace{0.2cm}{\bm{W}^T}\bm{W} = {\bm{I}_c},
\end{array}
\end{equation}
where $\bm{Y}^{(1)}$ and $\bm{Y}^{(2)}$ are Lagrange multiplier matrixes and $\mu  > 0$ is the penalty parameter.

\noindent 1) \textbf{Update $\bm{Z}$.} We solve the following problem to update $\bm{Z}$
\begin{equation}\begin{array}{cl}
{\bm{Z}_{t + 1}} = \hspace{-0.3cm}&{\mathop {\arg \min }\limits_{\bm{Z}} \left\| {\bm{X} - \bm{XZ}} \right\|_F^2 + \alpha \left\| \bm{Z} \right\|_F^2}\\[3mm]
&{ + \left\langle {\bm{Y}_t^{(1)},{\bm{W}_t} - {\bm{Z}^T}{\bm{P}_t}} \right\rangle}\\[3mm]
 &{+ \frac{{{\mu _t}}}{2}\left\| {{\bm{W}_t} - {\bm{Z}^T}{\bm{P}_t}} \right\|_F^2,}
\end{array}
\end{equation}
whose solution is given by
\begin{equation}{\bm{Z}_{t + 1}} = {({\bm{X}^T}\bm{X} + \lambda \bm{I} + \frac{\mu _t }{2}{\bm{P}_{t}}\bm{P}_{t}^T)^{ - 1}}{\bm{C}_{t}},
\end{equation}
where ${\bm{C}_{t}} = {\bm{X}^T}\bm{X} + {\bm{\bm{P}}_{t}}({\mu _t }{\bm{W}_t^{T}} + {\bm{Y}_t^{(1)T}})/2$.

\noindent 2) \textbf{Update $\bm{P}$.} We solve the following problem to update $\bm{P}$
\begin{equation}\begin{array}{cl}
{\bm{P}_{t + 1}} = &{\mathop {\arg \min }\limits_{\bm{P}} \frac{{{\mu _t}}}{2}(\left\| {{\bm{W}_t} - \bm{Z}_t^T\bm{P}} \right\|_F^2 + \left\| {{\bm{M}_t} - \bm{P}} \right\|_F^2) }\\[3mm]
 &{+ \left\langle {\bm{Y}_t^{(1)},{\bm{W}_t} - \bm{Z}_t^T\bm{P}} \right\rangle  + \left\langle {\bm{Y}_t^{(2)},{\bm{M}_t} - \bm{P}} \right\rangle,}
\end{array}
\end{equation}
whose solution is given by
\begin{equation}{\bm{P}_{t + 1}} = {({\bm{Z}_t}\bm{Z}_t^T + \bm{I})^{ - 1}}{\bm{D}_t},
\end{equation}
where ${\bm{D}_t} = {\bm{Z}_t}{\bm{W}_t} + {\bm{M}_t} + {{\bm{Z}_t}\bm{Y}_t^{(1)}}/{{{\mu _t}}} + {\bm{Y}_t^{(2)}}/{{{\mu _t}}}$.

\noindent 3) \textbf{Update $\bm{M}$.} We update $\bm{M}$ with the following problem
\begin{equation} {\bm{M}_{t + 1}} = \mathop {\arg \min }\limits_{\bm{M}} \left\langle {\bm{Y}_t^{(2)},\bm{M} - {\bm{P}_t}} \right\rangle  + \frac{{{\mu _t}}}{2}\left\| {\bm{M} - {\bm{P}_t}} \right\|_F^2,
\end{equation}
thus we obtain ${\bm{M}_{t + 1}} = {\bm{P}_t} - \bm{Y}_t^{(2)}/{\mu _t}.$

\noindent 4) \textbf{Update $\bm{W}$.} We update $\bm{W}$ with the following problem
\begin{equation}\begin{array}{cl}
{\bm{W}_{t + 1}} = &{\mathop {\arg \min }\limits_{\bm{W}} \left\langle {\bm{Y}_t^{(1)},\bm{W} - \bm{Z}_t^T{\bm{P}_t}} \right\rangle}\\[3mm]
 &{+ \lambda \hspace{0.05cm} tr({\bm{W}^T}{\bm{L}_t}\bm{W})
 + \frac{{{\mu _t}}}{2}\left\| {\bm{W} - \bm{Z}_t^T{\bm{P}_t}} \right\|_F^2}\\[3mm]
s.t.&{{\bm{W}^T}\bm{W} = {\bm{I}_c},}
\end{array}
\end{equation}
which is equivalent to the following problem:
\begin{equation}\begin{array}{cl}
{\bm{W}_{t + 1}} = &{\mathop {\arg \min }\limits_{\bm{W}} tr({\bm{W}^T}{\bm{L}_t}\bm{W}) + \eta \left\| {\bm{W} - {\bm{E}_t}} \right\|_F^2}\\[3mm]
s.t.&{{\bm{W}^T}\bm{W} = {\bm{I}_c},}
\end{array}
\end{equation}
where $\eta  = {\mu _t}/(2\lambda )$ and ${\bm{E}_t} = \bm{Z}_t^T{\bm{P}_t} - \bm{Y}_t^{(1)}/{\mu _t}$. First, let ${\bm{R}_t}\bm{R}_t^T = {\bm{L}_t} + \eta \bm{I}$, where $\bm{R}_t$ is a lower triangular matrix. Then, we perform the Singular Value Decomposition (SVD) on $\bm{E}_t^T{(\bm{R}_t^{ - 1})^T}{\bm{R}_t}$ and the result is denoted as $\bm{U}_t\bm{\Omega}_t {\bm{V}_t^{T}}$. Therefore, we get
\begin{equation}{\bm{W}_{t + 1}} = {\bm{V}_t}{\bm{I}_{c,n}}\bm{U}_t^T.
\end{equation}

\noindent 5) \textbf{Update $\bm{Y}^{(1)}$ and $\bm{Y}^{(2)}$.} For the Lagrange multipliers, we can update them as
\begin{equation}\begin{array}{l}
\bm{Y}_{t + 1}^{(1)} = \bm{Y}_t^{(1)} + {\mu _t}({\bm{W}_{t + 1}} - \bm{Z}_{t + 1}^T{\bm{P}_{t + 1}}),\\[3mm]
\bm{Y}_{t + 1}^{(2)} = \bm{Y}_t^{(2)} + {\mu _t}({\bm{M}_{t + 1}} - {\bm{P}_{t + 1}}).
\end{array}
\end{equation}

For clarity, the ADMM optimization algorithm for solving problem (\ref{admm}) is summarized in Algorithm \ref{Algorithm 1}.

\subsubsection{Fixing ${\bm{Z}_v}$ and ${\bm{P}_v}$, Update $\bm{A}$}
When ${\bm{Z}_v}$ and ${\bm{P}_v}$ are fixed, the model (11) can be formulated as
\begin{equation}\begin{array}{cl}
\mathop {\min }\limits_{\bm{A}}
&{\lambda \sum\limits_v {\sum\limits_{i,j} {\left\| {\bm{P}_v^T(\bm{z}_i^v - \bm{z}_j^v)} \right\|_2^2} {a_{ij}}} + \gamma \left\| \bm{A} \right\|_F^2}\\[3mm]
s.t.&{{\rm{  }}\forall i,{\rm{ }}\bm{a}_i^T\bm{1} = 1,{\rm{ }}{a_{ij}} \ge 0.}
\end{array}\label{15}
\end{equation}

For the problem (\ref{15}), due to the independence between different $i$, we can individually solve the following problem for each $i$ as
\begin{equation}\begin{array}{cl}
\mathop {\min }\limits_{{\bm{a}_i}} &{\lambda \sum\limits_v {\sum\limits_{j = 1}^n {\left\| {\bm{P}_v^T(\bm{z}_i^v - \bm{z}_j^v)} \right\|_2^2} {a_{ij}}}  + \gamma \sum\limits_{j = 1}^n {a_{ij}^2}} \\[3mm]
s.t.&{{\rm{   }}\bm{a}_i^T\bm{1} = 1,{\rm{ }}{a_{ij}} \ge 0,}
\end{array}
\end{equation}
which is equivalent to the following form
\begin{equation}\begin{array}{cl}
\mathop {\min }\limits_{{\bm{a}_i}} &{\sum\limits_{j = 1}^n {\left( {{d_{ij}}{a_{ij}} + \gamma a_{ij}^2} \right)}}\\[3mm]
s.t.&{{\rm{   }}\bm{a}_i^T\bm{1} = 1,{\rm{ }}{a_{ij}} \ge 0,}
\end{array}
\end{equation}
where ${d_{ij}} = \lambda \sum\nolimits_v {\left\| {\bm{P}_v^T\bm{z}_i^v - \bm{P}_v^T\bm{z}_j^v} \right\|_2^2}$. ${\bm{d}_i} \in {\Re ^{n \times 1}}$ is a vector, and the ${j^{th}}$ element in ${\bm{d}_i}$ is ${d_{ij}}$. Then, the above problem can be rewritten as
\begin{equation}\begin{array}{cl}
\mathop {\min }\limits_{\bm{a}{}_i} &{\left\| {{\bm{a}_i} + \frac{1}{{2\gamma }}{\bm{d}_i}} \right\|_2^2}\\[3mm]
s.t.&{{\rm{   }}\bm{a}_i^T\bm{1} = 1,{\rm{ }}{a_{ij}} \ge 0.}
\end{array}
\end{equation}

The Lagrangian function of the above problem is given by
\begin{equation}{\cal L}({\bm{a}_i},\Lambda ,{\bm{\varphi} _i}) = \frac{1}{2}\left\| {{\bm{a}_i} + \frac{1}{{2{\gamma _i}}}{\bm{d}_i}} \right\|_2^2 - \Lambda (\bm{a}_i^T\bm{1} - 1) - \bm{\varphi} _i^T{\bm{a}_i},
\end{equation}
where $\Lambda$ and ${\bm{\varphi} _i}$ are Lagrangian multipliers. For each data point $\bm{x}_i$, we set the number of the nearest neighbours as $k$. So we can obtain the optimal solution of $\bm{a}_i$ according to Karush-Kuhn-Tucker (KKT) condition
\begin{equation}{a_{ij}} = {( { - \frac{{{d_{ij}}}}{{2{\gamma _i}}} + \frac{1}{k} + \frac{1}{{2k{\gamma _i}}}\sum\limits_{j = 1}^k {{d_{ij}}} } )_ + }.\label{29}
\end{equation}

In the formula (\ref{29}), we make ${d_{i1}}$, ${d_{i2}}$,..., ${d_{in}}$ be sorted in the ascending order. Then, to let most elements of $\bm{a}_i$ have exact $k$ non-zero elements, we have
\begin{equation}{\gamma _i} = \frac{k}{2}{d_{i,k + 1}} - \frac{1}{2}\sum\limits_{j = 1}^k {{d_{ij}}} .
\end{equation}

Hence, we can determine the final parameter $\gamma$ by computing the average of $\gamma _i$ \cite{nie2017multi}
\begin{equation}\gamma  = \frac{1}{n}\sum\limits_{i = 1}^n {{\gamma _i}}  = \frac{1}{n}\sum\limits_{i = 1}^n {(\frac{k}{2}{d_{i,k + 1}} - \frac{1}{2}\sum\limits_{j = 1}^k {{d_{ij}}} )} .
\end{equation}

For clarity, the whole optimization procedure for solving
problem (\ref{11}) is summarized in Algorithm \ref{Algorithm 2}.

\subsection{Convergence Analysis}
The original problem (\ref{11}) is not jointly convex and can be divided into two optimization sub-problems. Each of them is the convex minimization problem and we can obtain the optimum solution for each sub-problem. Therefore, the original objective function is non-increasing with the iterations until Algorithm \ref{Algorithm 2} converges.

\section{Experiments}

In this section, we evaluate our proposed MSCAN and MSCAM methods on a synthetic dataset and three real-world benchmark datasets to demonstrate their effectiveness.

\subsection{Dataset Descriptions}
 The descriptions of the datasets used in our experiments are summarized in Table \ref{2}.

\noindent \textbf{Synthetic dataset} is constructed according to \cite{gao2016incomplete}. Specifically, this synthetic dataset is comprised of three views with two clusters. Each view is generated from a two-component Gaussian mixture model and 1,000 data points are sampled as instances in each view. Finally, we use noises to uniformly corrupt the data points with the percentage $10 - 90\% $ of entries at random, e.g., we choose a column $\bm{x}$ to corrupt by adding the Gaussian noise (zero mean and variance $0.3{\left\| \bm{x} \right\|_2}$) to its observed vector.

\noindent \textbf{Oxford Flowers dataset} is composed of 1,360 examples that consist of 17 flower categories. Each category is composed of 80 images. Different features (color, texture, shape) are used to describe each image. In addition, ${\chi ^2}$ distance matrixes for three different visual features (color, texture, shape) are utilized to construct three views.

\noindent \textbf{Handwritten numerals (HW) dataset} contains 2,000 examples from 0 to 9 digit classes. Each class has 200 examples. We use six public features including 240-dimension pixel averages in $2 \times 3$ windows (PIX), 216-dimension profile correlations (FAC), 76-dimension Fourier coefficients of character shapes (FOU), 47-dimension Zernike moment (ZER), 64-dimension Karhunen-love coefficients (KAR) and 6-dimension morphological (MOR) features.

\noindent \textbf{NUS-WIDE-Object (NUS) dataset} is a web image dataset for object recognition. It consists of 30,000 images in 31 categories. In our experiments, we randomly extract 100 images for each category. We use six low-level features to describe the image: 64-dimension color histogram (CH), 225-dimension block-wise color moments (CM), 144-dimension color correlation (CORR), 73-dimension edge direction histogram (EDH), 128-dimension wavelet texture (WT) and 500-dimension BoW SIFT.

\subsection{Experiment Setup} We evaluate MSCAN and MSCAM by comparing with other state-of-the-art clustering methods. Specifically, we compare with the single-view clustering methods: Spectral Clustering (SC) \cite{ng2002spectral} and Sparse Subspace Clustering (SSC) \cite{elhamifar2013sparse}. The multi-view clustering methods: Co-regularized Spectral Clustering (CoReg) \cite{kumar2011co}, Multi-Modal Spectral Clustering (MMSC) \cite{cai2011heterogeneous}, Diversity-induced Multi-view Subspace Clustering (DiMSC) \cite{cao2015diversity}, and Multi-view Learning
with Adaptive Neighbours (MLAN) \cite{nie2017multi}, are also utilized as comparison methods. The detailed information of these comparison approaches is described as follows.
\begin{itemize}
\item \textbf{SC-BSV:} SC is a classic spectral clustering method. In our experiments, we perform the SC method on each single-view feature and report the best results.
\item \textbf{SSC-BSV:} SSC is a representative subspace clustering method based on self-representation. In this method, subspace representation is obtained at first, and then spectral clustering method is performed on the subspace representation. In our experiments, we employ the SSC method for each single view and report the best results.
\item \textbf{SC-Concat:} We first concatenate all features into a long vector and then perform SC to get final clustering results.
\item \textbf{SSC-Concat:} Same as SC-Concat, we perform SSC to get the final clustering results of concatenated features.
\item \textbf{CoReg:} This method introduces a centroid-based co-regularization term to make all of views have the same clustering results.
\item \textbf{MMSC:} This method learns a shared Laplacian matrix by integrating multi-view heterogeneous image features. In addition, a non-negative constraint is utilized to improve robustness of this model.
\item \textbf{DiMSC:} This method utilizes the Hilbert Schmidt Independence Criterion (HSIC) as a diversity term to explore the complementarity information of multi-view data.
\item \textbf{MLAN:} This approach simultaneously performs multi-view clustering and local structure learning. Moreover, the weight for each view is automatically determined without additional penalty parameters.
\end{itemize}
\begin{table}[t]
\scriptsize
\renewcommand{\arraystretch}{1.05}
\centering
\caption{Datasets used in our experiments.}\label{2}
\resizebox{0.45\textwidth}{!}{
\begin{tabular}{|c|c|c|c|}\hline
Datasets&\#.Size&\#.View&\#.Cluster\\\hline
Synthetic Dataset&1000&3&2\\
Oxford Flowers&1360&3&17\\
Handwritten&2000&6&10\\
NUS-WIDE-Object&3100&6&31\\
\hline
\end{tabular}}\vspace{-0.1cm}
\end{table}

There are three trade-off parameters $\alpha$, $\lambda$, and $\gamma$ in our model, where the parameter $\gamma$ can be determined according to the property of adaptive neighbours \cite{nie2017multi}. Therefore, to obtain the best clustering results, we only need to adjust parameters $\alpha$ and $\lambda$. In addition, for each data point, the number of the nearest neighbours is set as 9. For all compared methods, we also tune their parameters to obtain the best results. Besides, we conduct experiments for 10 times to get the average results for each dataset. All experiments are performed by MATLAB tool on computer with Intel Xeon E5-2650 v2CPU (2.6GHz) and 32G RAM.
\begin{table*}[t]
\scriptsize
\renewcommand{\arraystretch}{1.05}
\centering
\caption{Clustering ACC on synthetic dataset. The best results are in bold font.}\label{1}
\resizebox{0.9\textwidth}{!}{
\begin{tabular}{|c|c|c|c|c|c|c|c|c|c|c|}\hline
Corruptions (\%)&0&10&20&30&40&50&60&70&80&90\\\hline
\hline
MLAN&0.849&0.841&0.833&0.830&0.787&0.732&0.683&0.668&0.600&0.579\\
\hline
\hline
\textbf{MSCAN}&\textbf{0.890}&0.841&0.836&0.832&0.813&0.805&0.763&0.758&0.748&0.702\\
\textbf{MSCAM}&0.874&\textbf{0.872}&\textbf{0.868}&\textbf{0.846}&\textbf{0.840}&\textbf{0.833}&\textbf{0.826}&\textbf{0.817}&\textbf{0.788}&\textbf{0.780}\\
\hline
\end{tabular}}
\end{table*}

\subsection{Experiment Results on Synthetic Dataset}
We compare MSCAN and MSCAM with the recently proposed MLAN \cite{nie2017multi} that learns the similarity matrix with original data.
Experimental results (accuracy) are shown in Table \ref{1}. It is noteworthy that the proposed MSCAN and MSCAM methods consistently outperform MLAN method. In addition, when the noise level is high (percentage is 90\%), the clustering result of MLAN is very bad, i.e., 57.9\%. In contrast, MSCAN and MSCAM can still achieve promising clustering performance, i.e., 70.2\% and 78.0\%. Further, MSCAN has a better clustering result than MSCAM on the original synthetic dataset. However, MSCAM consistently outperforms MSCAN on the corrupted synthetic dataset. This indicates that the MSCAM method is more robust than MSCAN and MLAN.
\begin{table}[!t]
\renewcommand{\arraystretch}{1.2}
\centering
\caption{Clustering ACC (mean and standard deviation) of different methods. The best results are in bold font.}\label{tablep1}\vspace{0.1cm}
\resizebox{0.45\textwidth}{!}{
\begin{tabular}{|c|c|c|c|}\hline
Dataset&Oxford Flowers&HW&NUS\\\hline
\hline
SC-BSV&0.411(0.003)&0.723(0.000)&0.131(0.002)\\
SSC-BSV&0.356(0.008)&0.767(0.004)&0.147(0.001)\\
SC-Concat&0.428(0.012)&0.752(0.000)&0.142(0.002)\\
SSC-Concat&0.365(0.006)&0.815(0.009)&0.152(0.001)\\
CoReg&0.433(0.011)&0.804(0.059)&0.188(0.007)\\
MMSC&0.442(0.013)&0.840(0.011)&0.154(0.005)\\
DiMSC&0.431(0.021)&0.907(0.004)&0.109(0.003)\\
MLAN&0.459(0.001)&0.973(0.000)&0.104(0.002)\\
\hline
\hline
\textbf{MSCAN}&0.527(0.004)&0.975(0.000)&0.179(0.005)\\
\textbf{MSCAM}&\textbf{0.530(0.004)}&\textbf{0.978(0.001)}&\textbf{0.190(0.002)}\\
\hline
\end{tabular}}
\end{table}

\begin{table}[!t]
\renewcommand{\arraystretch}{1.2}
\centering
\caption{Clustering NMI (mean and standard deviation) of different methods. The best results are in bold font.}\label{tablep2}\vspace{0.1cm}
\resizebox{0.45\textwidth}{!}{
\begin{tabular}{|c|c|c|c|}\hline
Dataset&Oxford Flowers&HW&NUS\\\hline
\hline
SC-BSV&0.426(0.005)&0.667(0.000)&0.042(0.003)\\
SSC-BSV&0.373(0.005)&0.759(0.001)&0.013(0.000)\\
SC-Concat&0.434(0.010)&0.709(0.000)&0.089(0.002)\\
SSC-Concat&0.410(0.005)&0.850(0.007)&0.031(0.001)\\
CoReg&0.423(0.008)&0.778(0.035)&0.209(0.006)\\
MMSC&0.446(0.004)&0.892(0.008)&0.172(0.004)\\
DiMSC&0.443(0.011)&0.841(0.003)&0.102(0.004)\\
MLAN&0.476(0.002)&0.939(0.001)&0.091(0.003)\\
\hline
\hline
\textbf{MSCAN}&\textbf{0.540(0.002)}&0.942(0.000)&0.206(0.003)\\
\textbf{MSCAM}&0.522(0.002)&\textbf{0.948(0.001)}&\textbf{0.214(0.002)}\\
\hline
\end{tabular}}
\end{table}
\subsection{Experiment Results on Benchmark Datasets}
In our experiments, we utilize accuracy (ACC) and normalized mutual information (NMI) as two evaluation metrics to evaluate clustering methods. Table \ref{tablep1} and Table \ref{tablep2} show the clustering results on three benchmark datasets, respectively. In general, the multi-view clustering methods can achieve superior results than single-view approaches (SC and SSC). Additionally, we can observe that the proposed MSCAM method achieves the best clustering performance in comparison with other state-of-the-art multi-view clustering methods (CoReg, MMSC, DiMSC, and MLAN).

For HW dataset, the recently proposed MLAN method achieves high performance, so the improvement of our MSCAM method is not apparent.
However, for the large NUS dataset, MSCAM significantly improves the clustering performance compared with MLAN. This is because that MLAN constructs the graph by utilizing original data that have noises and outliers for a large dataset. By contrast, MSCAM obtains a better graph by learning different subspace representations and Mahalanobis matrixes for different views, and thus achieves a better clustering performance.

Additionally, to better evaluate MSCAM, we further report the clustering performance of MSCAN and MSCAM. We can observe that MSCAN can achieve competitive or even better clustering results than MSCAM for small-scale datasets (Oxford Flowers and HW). However, due to the complexity of noises and outliers in the large NUS dataset, MSCAM outperforms MSCAN. This is because that the unchangeable similarity relationships among subspace representations result in the suboptimal performance in MSCAN. MSCAM learns different Mahalanobis matrixes for all views, which can change the similarity among subspace representations, and further obtains a better result. Consequently, the proposed MSCAM is robust and can achieve superior performance than other clustering algorithms.

\begin{figure}[!t]
\begin{minipage}[t]{0.49\linewidth}
\centering
\includegraphics[width=1.58in]{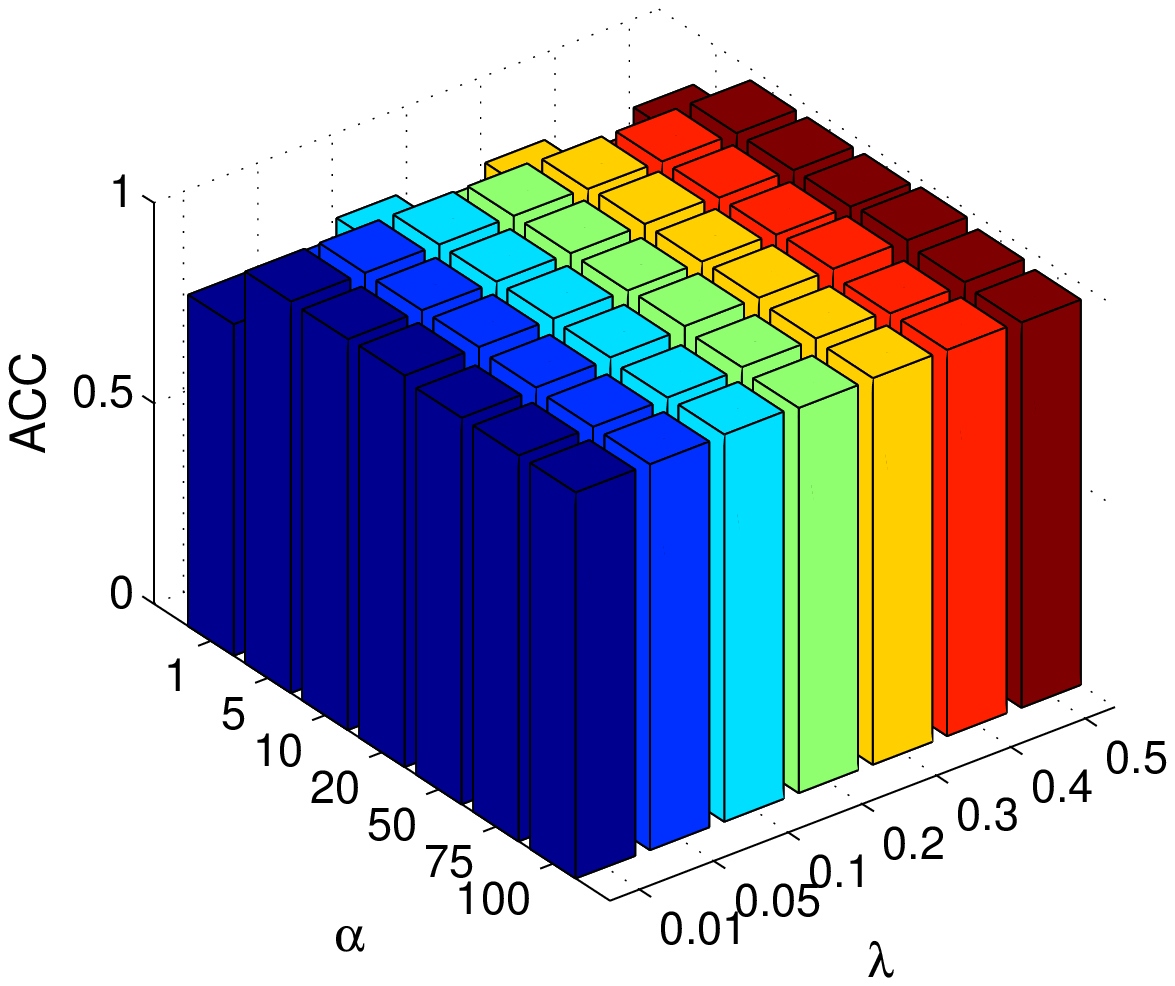}
\end{minipage}
\begin{minipage}[t]{0.49\linewidth}
\centering
\includegraphics[width=1.58in]{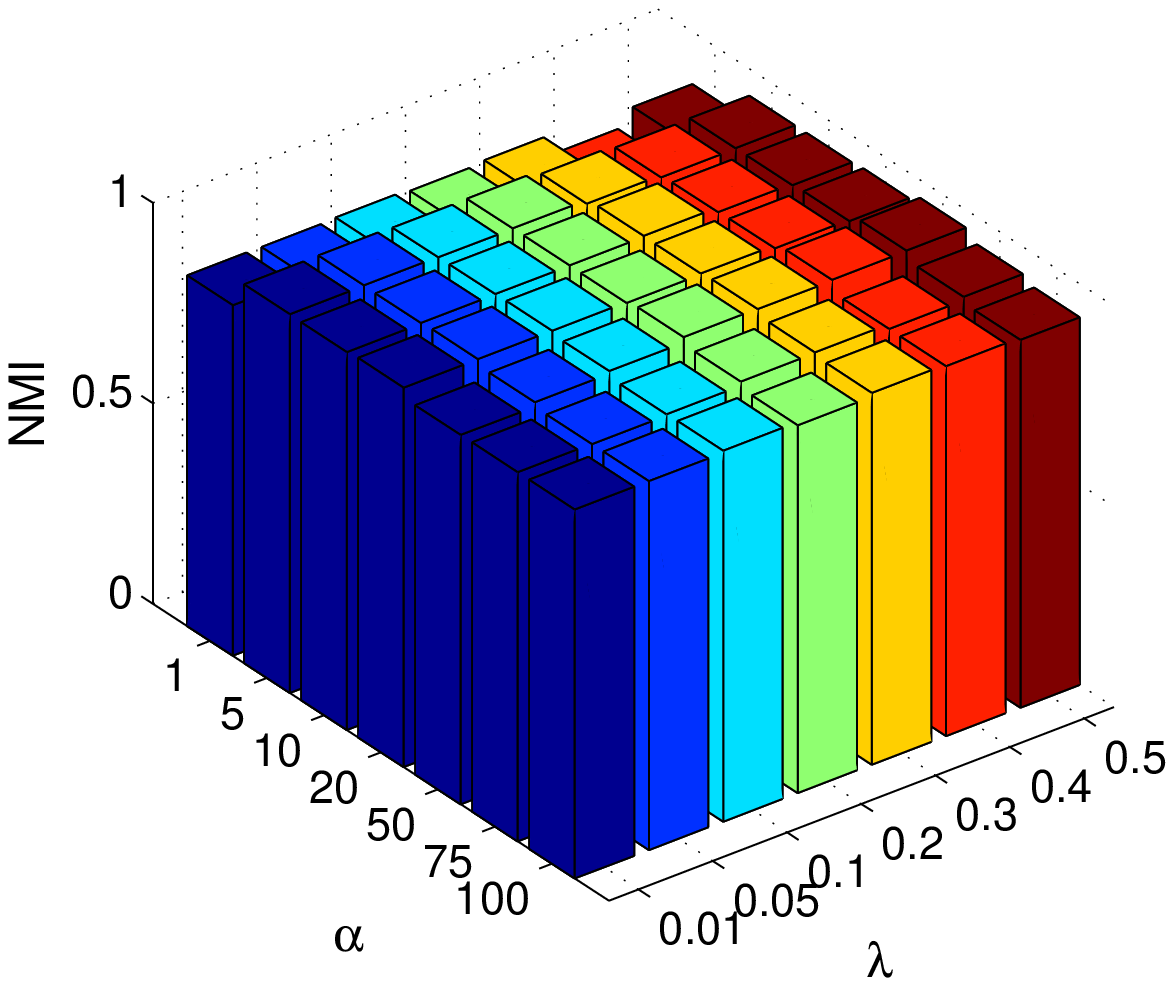}
\end{minipage}\vspace{-0.1cm}
\caption{Clustering results on HW dataset w.r.t. $\alpha$ and $\lambda $.}
\label{con}\vspace{-0.1cm}
\end{figure}

\subsection{Parameter Sensitivity}
In our method, there are two parameters $\alpha$ and $\lambda$. Figure \ref{con} shows the parameter sensitivity of MSCAM on HW dataset. It is obvious that our proposed method is not very sensitive to parameters and can achieve satisfactory clustering performance within a large range of parameter values ($\alpha$ and $\lambda$).

\section{Conclusion}
In this paper, we propose a novel multi-view subspace clustering method MSCAM which joints adaptive neighbours and metric learning. Our method learns subspace representations of original data and uses them to adaptively learn a consensus similarity matrix. Meanwhile, considering different contributions of views, we utilize Mahalanobis metric to learn different projections. Further, we add a graph constraint to remove the \emph{k}-means procedure. We develop an iterative optimization algorithm for MSCAM. Extensive experimental results on a synthetic dataset and three real-world datasets demonstrate that our MSCAM is robust and outperforms other state-of-the-art clustering algorithms.


\end{document}